\documentclass{opt2022} % Include author names
\usepackage{nicefrac}
\usepackage{microtype}      % microtypography
\usepackage{algorithm}
\usepackage[shortlabels]{enumitem}
\usepackage{algorithmic}
\usepackage{amsmath,amsfonts,bm}

\allowdisplaybreaks

\usepackage{mathtools}
\usepackage{cancel}

\def\gD{{\mathcal{D}}}

\def\gX{{\mathcal{X}}}

\newtheorem{assumption}{Assumption}
\def\sR{{\mathbb{R}}}

\newcommand{\E}{\mathbb{E}}

\DeclareMathOperator*{\argmin}{arg\,min}

\newcommand{\PS}{\mathrm{PS}}

\newcommand{\PR}{\mathrm{PR}}
\newcommand{\DPR}{\mathrm{DPR}}
\newcommand{\thetaPO}{{\theta_\mathrm{PO}}}
\newcommand{\thetaPS}{{\theta_\mathrm{PS}}}

\title[Optimizing the Performative Risk under Weak Convexity Assumptions]{Optimizing the Performative Risk under \\ Weak Convexity Assumptions}

\optauthor{%
\Name{Yulai Zhao} \Email{yulaiz@princeton.edu}\\
\addr Princeton University
}

\begin{document}

\maketitle
\label{sec:abs}

\begin{abstract}
In performative prediction, a predictive model impacts the distribution that generates future data, a phenomenon that is being ignored in classical supervised learning. 
In this closed-loop setting, the natural measure of performance named performative risk ($\mathrm{PR}$), captures the expected loss incurred by a predictive model \emph{after} deployment.  
The core difficulty of using the performative risk as an optimization objective is that the data distribution itself depends on the model parameters.
This dependence is governed by the environment and not under the control of the learner. 
As a consequence, even the choice of a convex loss function can result in a highly non-convex $\mathrm{PR}$ minimization problem. 
Prior work has identified a pair of general conditions on the loss and the mapping from model parameters to distributions that implies the convexity of the performative risk. 
In this paper, we relax these assumptions and focus on obtaining weaker notions of convexity, without sacrificing the amenability of the $\mathrm{PR}$ minimization problem for iterative optimization methods.
\end{abstract}

\section{Introduction}
\label{sec:intro}

% \cnote{I would like to rework the intro a bit, but we should first converge on the results and message in the main sections 3/4. }

Predictions in the real world are often performative. This means predictions because they are made public, or because they inform downstream actions, and influence the distribution of the features or the target variable they aim to predict. Thus, shifting the data distribution the model has been trained on. Prominent examples include stock price prediction, credit scoring, and predictive policing.

As a result, the performance of a model should not be measured with respect to a fixed underlying distribution, as in classical supervised learning,  but instead on the very distribution, the model induces.  This has been conceptualized by~\cite{perdomo2020performative} through the framework of \emph{performative prediction}.  As a dynamic setting gives rise to interesting and new optimization challenges, the objective function in performative prediction is the risk incurred by the learner on the distribution induced after deploying the model:
\begin{gather}
\label{eq:PR}
     \PR(\theta)\coloneqq \E_{z \sim \gD(\theta)} \ell(z; \theta), \theta \in \Theta
\end{gather}
where $\Theta$ is a closed and convex set in $\sR^d$. And $\PR$ is referred to as the performative risk and measures the loss of a predictor $\theta$ on the distribution $\gD(\theta)$ resulting from its deployment. We focus on finding the minima of the performative risk (\ref{eq:PR}), known as performative optimal points.
The main challenge of performative risk minimization from an optimization perspective is that the dependence of the performative risk on the model parameters is two-fold: it appears both in the distribution $\gD$ and the loss $\ell$. To decouple these two dependencies, let us define the \emph{decoupled performative risk}
\begin{align*}
    \DPR(\theta_1, \theta_2) \coloneqq \E_{z \sim \gD(\theta_1)} \ell(z; \theta_2)
\end{align*}
to denote the loss of $\theta_2$ measured over the distribution $\gD(\theta_1)$. According to this definition we have $\PR(\theta) = \DPR(\theta, \theta).$ 

Optimizing the performative risk directly is challenging because the structure of the objective $\PR(\theta) = \E_{z \sim \gD(\theta)} \ell(z; \theta)$ is not under the control of the learner and can be highly non-convex. Thus, for tackling performative risk minimization by building on classical tools from the optimization literature we need to better understand the structure of the objective function, and how properties of the distribution map and the loss interact.

As an important first step, \citet{miller2021outside} identified a natural set of properties of the loss function and model-induced
distribution shift under which the performative risk is convex.
In particular, they showed that linear dependence of the distribution map on the model parameters $\theta$, together with strong convexity of the loss function $\ell$ implies global convexity of $\PR$. In this work, we are interested in studying problems beyond this specific setting and we intend to address the following question:

\begin{center}
    \emph{How and under what conditions can we optimize the performative risk?}
\end{center}
In particular, we study how to optimize the performative risk through first-order information and analyze the suboptimality gap of performative risks.
Noticing the fact that convexity can be relaxed to
weaker conditions while the same convergence guarantees are preserved by gradient-based
optimizer, in Section~\ref{sec:ext}, we show weaker conditions on $\DPR$ still imply desirable local properties of $\PR$. The properties enjoy generality since they do not rely on global structural restrictions. In Section~\ref{sec:compare}, we discuss what information about $\thetaPO$ we could have, if starting from a stable point. The underlying idea is to make use of previous retraining methods~\citep{perdomo2020performative,mendler2020stochastic} converging to $\thetaPS$ while the ultimate goal is still reaching $\thetaPO$.

\section{Background}
\label{sec:back}

Performativity is a phenomenon that is widely recognized in finance~\citep{clarke2012financial}, economics~\citep{callon2007does,cochoy2010performativity} and other social sciences~\citep{roberts2009performative}. Following the seminal work by \cite{perdomo2020performative}  performativity has increasingly gained attention from the machine learning community. Performative distribution shifts constitute an instance of a more general distribution shift problem, where the shift is endogenous to the problem and entirely caused by the deployed predictive model. 
%Particularly, \citet{perdomo2020performative} highlighted that predictive models, when deployed in the social world, can cause performative feedback effects: traffic predictions impact route choices, credit scoring rules impact consumption patterns, and predictions on whether a student will pass an exam, directly impact their scholastic performance. 
In other words, the deployment of a predictive model represents an intervention to the population~\citep{mendler22}. The resulting causal feedback effects invalidate the standard assumption of supervised learning where there is a static and model-independent data distribution describing the population, surfacing as distribution shifts.

%including reinforcement learning~\citep{perdomo2020performative}, causal inference~\citep{pearl2009causality}, and the application of algorithmic decision making in social contexts more broadly~\citep{cochoy2010performativity}. 

%Taking performativity into account when training machine learning models gives rise to many interesting and new challenges for optimization and algorithm design. 

%In the following we will first review the performative prediction framework that allows us to conceptualize this phenomenon, relevant algorithmic papers, and introduce relevant concepts.

% \subsection{Performative prediction framework}

\subsection{Key concepts}
We first give an overview of two existing concepts characterizing whether $\theta$ is a desirable solution of performative prediction.

\paragraph{Performative stability.} Performative prediction can be viewed as a game between the decision-maker and the population that responds to the deployed model. The first concept of optimality in this context is performative stability 
\begin{equation} 
\label{eq:stable}
    \thetaPS=\argmin_\theta \DPR(\thetaPS, \theta).
\end{equation}

The expected loss over $\gD(\theta_\PS)$ is minimized at $\theta_\PS$. As such stable points constitute a fixed point of so-called \emph{retraining} methods, which are heuristically adopted in dealing with distribution shifts. More specifically, at the $t$-th iteration, these methods find a set of parameters that minimizes the risk on the previously-generated distribution $\gD(\theta_t)$. It can be formally described as $
\theta_{t+1} = \arg \min_{\theta \in \Theta}  \DPR(\theta_t, \theta).
$
Several works have studied the convergence of population-based and stochastic retraining methods to stable points in the standard framework~\citep{perdomo2020performative, mendler2020stochastic, drusvyatskiy2020stochastic} as well as state-dependent extensions~\citep{brown2020performative,li2021state, harris2021stateful}.

\paragraph{Performative optimality.} An alternative solution concept is performative optimality. Performative optimal points describe models that achieve the minimal loss on the distribution they induce. In this sense, they are globally optimal in the following sense:
\[\thetaPO=\argmin_{\theta \in \Theta} \DPR(\theta,\theta).\]
It follows from the definition that $\PR(\thetaPO)\leq\PR(\thetaPS)$ while in general, stable points are not necessarily optimal and optimal points are not necessarily stable. Furthermore, the two solutions $\thetaPS$ and $\thetaPO$ can be arbitrarily far apart from each other, see~\citep{perdomo2020performative, miller2021outside}. Thus, retraining does not, in general, serve as a reliable heuristic for finding optimal points. That being said, we focus on finding performative optima in this work. There is only a handful of works that study approaches for finding optimal points. The first to tackle performative optimality was~\citet{miller2021outside}. They proposed an iterative optimization method for minimizing the performative risk directly by first learning a model of the distribution map and then applying a gradient-based procedure to the resulting estimation of the performative risk. To guarantee the global convergence of such an approach they studied structural assumptions under which the performative risk is convex. Around the same time, ~\citet{izzo2021learn} studied \emph{performative gradient descent}, an algorithm that directly estimates the gradient of the performative risk. Technically, they assumed the existence of an estimator $\hat{f}(\theta)$ so that one can use samples from the distribution $\gD(\theta)$ to infer $\theta$, an assumption we will revisit later in this manuscript. To prove convergence to optimal points they directly assume convexity of the performative risk, an assumption that is hard to verify.  

A different approach has recently been taken by~\citet{jagadeesan2022regret} who leveraged tools from the bandits' literature for minimizing regret under performativity without making structural assumptions on the performative risk apart from weak regularity assumptions. Technically, they use data collected from experiments to build performative confidence bounds to characterize the regret of unexplored parameters. % Comparing these work, the model-based algorithm from~\citep{miller2021outside} does not assume knowledge of the base distribution $\gD(0)$ while ~\citet{jagadeesan2022regret} does. Moreover, theoretical guarantees from~\citep{miller2021outside} can be easily translated into a regret bound that resembles ~\citep[Theorem 4]{jagadeesan2022regret}. However, the latter theorem applies to a more general setting, requiring fewer restrictions on the loss function and PR itself.
Their method achieves sublinear regret and is the first to find optimal points for non-convex performative risk functions. However, a significant drawback of such a bandit approach is the need for global exploration, i.e., one needs to explore the entire parameter space to find the minimizer, something that is often not feasible in practical applications. Therefore, the focus of this work is on iterative descent methods that gradually approach optimal points.

% \paragraph{Strategic classification as a special case} 
% Closely related to performative prediction is the long line of work on strategic classification initiated by~\citet{dalvi2004adversarial,hardt2016strategic} that studies the interactions between a classification rule and strategic agents that aim to achieve a favorable classification by manipulating their features. As a special case of performative prediction, strategic classification imposes a specific structure on the distribution map. 
% Many works on strategic classification have studied finding performative optima.~\citep{milli2019social, chen2020learning, shavit2020causal, bechavod2021gaming}

\subsection{Assumptions on the loss function}

The loss function in performative prediction takes the same role as the loss in supervised learning. We discuss classical assumptions from the optimization literature such as strong convexity and smoothness on the loss function.

\begin{assumption}[Smoothness] \label{assump_smooth}
The loss $\ell(\cdot; \cdot)$ is $\beta$-smooth if for all $z, z^\prime, \theta$,
    \begin{equation*}
        \|\nabla_\theta \ell(z; \theta) -\nabla_\theta \ell(z^\prime; \theta) \|_2 \le \beta \|z - z^\prime\|_2.
    \end{equation*}
\end{assumption}

\begin{assumption}[Strong convexity]
\label{assump_strong_cvx}
The loss $\ell(\cdot; \cdot)$ is $\gamma$-strongly convex in $\theta$ if for all $\theta, \theta^\prime, z$,
    \begin{equation*}
        \ell(z; \theta^\prime) - \ell(z; \theta) \ge \nabla_\theta \ell(z; \theta)^\top (\theta^\prime - \theta) + \frac{\gamma}{2} \|\theta - \theta^\prime\|_2^2.
    \end{equation*}
Note that this condition reduces to convexity when $\gamma=0$.
\end{assumption}

These two assumptions characterize the dependence of the loss function on the model parameters $\theta$ and they are assumed to hold for any realization of $z$. We will adopt the smoothness assumption in this work and we will discuss the strong-convexity assumption as well as weaker notions of convexity.

In addition, we will need a regularity assumption for relating changes in the data distribution to changes in the loss function. Therefore, we will adopt the Lipschitzness condition w.r.t. $z$, which is common when dealing with performative distribution shifts~\citep{perdomo2020performative,miller2021outside}, namely

\begin{assumption}[Lipschitzness] \label{assump_lipschitz}
The loss $\ell(\cdot; \cdot)$ is $L$-Lipschitz in $z$ if for all $z, z^\prime, \theta$,
    \begin{equation*}
        \|\nabla_z \ell(z; \theta) -\nabla_z \ell(z^\prime; \theta) \|_2 \le L \|z - z^\prime\|_2.
    \end{equation*}
\end{assumption}
\subsection{Characterizing the performative shift}
\label{sec:performative}

Since performative effects are not under the control of the learner and are typically unknown, we need assumptions on these distribution shifts for studying the problem and establishing convergence guarantees of algorithmic approaches. In contrast to classical supervised learning where optimization techniques are commonly adopted under conditions on the loss function, we need additional assumptions for performative optimization. This observation is at the core of the main difficulty in performative prediction and the art is to couple conditions on the loss function with conditions on the distribution shift to obtain favorable properties of the resulting performative risk minimization problem.

We shall first introduce the concept of Wasserstein distance, then we proceed to the key sensitivity assumption that relates changes in the model parameters to changes in the distribution.

\begin{definition}[Wasserstein Distance]
The Wasserstein-1 distance, also called Earth-Mover (EM) distance, is defined as
\begin{equation} \label{eq::W}
W(P_r, P_g) = \inf_{\gamma \in \Pi(P_r ,P_g)} \E_{(x, y) \sim \gamma}\big[\:\|x - y\|\:\big]~,
\end{equation}
where $\Pi(P_r,P_g)$ denotes the set of all joint distributions $\gamma(x,y)$ whose marginals are respectively $P_r$ and $P_g$. 
\end{definition}
Intuitively, $\gamma(x,y)$ indicates how much mass must be transported from $x$ to $y$ in order to transform the distributions $P_r$ into the distribution $P_g$. The EM distance then is the "cost" of the optimal transport plan. The infimum in Eq.~\ref{eq::W} is highly intractable. On the other hand,
the Kantorovich-Rubinstein duality tells us that
\begin{equation*}
W(P_r, P_\theta) = \sup_{\|f\|_L \leq 1} \left[ \int f(x) d P_r(x) - \int f(x) d P_\theta (x) \right]
\end{equation*}
This allows us to relate the distance of the expectation of a Lipschitz function over two different distributions by their Wasserstein distance. 

\paragraph{Sensitivity.}
One commonly used assumption is a weak regularity condition that posits Lipschitzness
of the distribution shift relative to changes in the model parameters.
Intuitively, the idea is, if decisions are made according to similar predictive models, then the resulting distributions should also be close. 
More precisely, the \emph{sensitivity condition} introduced by \cite{perdomo2020performative} relates changes in the model parameter as measured in Euclidean distance to changes in the distribution as measured in Wasserstein distance. 
\begin{assumption}[Sensitivity]
\label{assump_sensitive}
The distribution map $\gD(\cdot)$ is $\epsilon$-sensitive if for all $\theta, \theta^\prime$,
    \begin{equation*}
        W (\gD(\theta), \gD(\theta^\prime)) \le \epsilon \|\theta - \theta^\prime \|_2.
    \end{equation*}
\end{assumption}

A prominent example of a distribution shift that satisfies the sensitivity assumption is strategic classification with quadratic costs~\citep{perdomo2020performative}. Strategic classification microfounds performative prediction with the best response model for individual actions~\citep{ jagadeesan2021alternative}. 

\paragraph{Mixture dominance.} Below we present a stronger and more structural assumption on the distribution shift introduced by \citet{miller2021outside} which guarantees the convexity of $\DPR(\theta^\prime, \theta)$ w.r.t. the first argument $\theta^\prime$. 

\begin{assumption}[Mixture dominance]
\label{assump_mix}
A set of losses and functions satisfy the mixture dominance condition if $\DPR(\theta^\prime, \theta)$ is convex w.r.t. $\theta^\prime$, thus for all $\theta, \theta^\prime$,
\begin{equation*}
        \E_{z \sim \gD(\theta)} [\ell(z; \theta) \nabla_\theta \log{(p_\theta(z))}]^\top (\theta^\prime - \theta) \le \DPR(\theta^\prime, \theta) - \PR(\theta).
    \end{equation*}
\end{assumption}

 It has been demonstrated by~\cite{miller2021outside} that this assumption is satisfied in performative environments by the well-known \emph{location-scale} distribution family, which is a family of distributions formed by translation and re-scaling of a standard family member. For example, let the underlying distribution be $\gD_\theta$, data points $z_\theta \sim \gD_\theta$ satisfies $z_\theta = z_0 + A \theta$, where the matrix $A$ is an unknown parameter, and $z_0$ is is a zero-mean random variable. Here the additive term $A \theta$ represents the effects of performativity. Standard examples include Gaussian, Cauchy and uniform distributions. There have been wide applications for these families in economic~\citep{wong2008preferences} and statistics~\citep{hazra2017stochastic}. Examples of location families can also be found in strategic classification~\citep{hardt2016strategic}.

\section{Beyond convexity of the performative risk}
\label{sec:ext}

In studying the convergence of gradient-based methods to performative optima, previous works on performative prediction have focused on a setting where the performative risk is convex. 
While some works such as ~\cite{izzo2021learn} directly start from apriori hard-to-verify assumptions, the notable work by \citet{miller2021outside} provides conditions on the loss function and the distribution map under which global convexity of $\PR$ emerges. 
More specifically they show that strong convexity of the loss function, together with 
Assumptions~\ref{assump_sensitive}-\ref{assump_mix} to control the performativity effects of distributions, is sufficient to establish (strong) convexity of the performative risk.  They obtain a zeroth-order algorithm that provably converges towards globally performative optimal points ($\thetaPO$)~\citep{miller2021outside}, which could avoid generally difficult computing performative gradients. 

%Then, given convexity, \citet{miller2021outside} proposed a way to optimize PR directly and proved \emph{strong-convexity} of PR for this goal \cnote{are you sure? do their results not apply for $\mu=0$?}. Apart from the essential Assumptions~\ref{assump_sensitive}, ~\ref{assump_mix} to control the performativity effects of distributions, they further require a strong Assumption~\ref{assump_strong_cvx} of $\ell(\cdot;\cdot)$ w.r.t. $\theta$ and $z$. 

In the following, we take a step further by demonstrating that the strong convexity assumption on the loss function can be relaxed while preserving the amenability of the performative risk minimization problem to gradient-based optimization. We build on the existing literature on first-order optimization that showed that convexity can be relaxed to weaker conditions while maintaining the same convergence guarantees of gradient-based optimizers.
%rely on the fact that numerous conditions have been explored in first-order optimization to show convergence beyond (strong) convexity.

%linear convergence rates without strong convexity~\citep{karimi2016linear}. \cnote{do you mean convexity? we care about PR that is not convex} This raises the question whether
%PR could be shown to satisfy these conditions (e.g., EB, PL, WC), which could relax the requirements needed to prove convergence to optimal points. To address this problem, we explore how assumptions on the loss translate to properties of the performative risk.

\subsection{Weaker notions of strong convexity}
\label{sec:convexity}
In the optimization literature, strongly-convex optimization objectives are playing an important role as they can be optimized using gradient-descent methods at a linear rate. In the context of finding performatively stable points, this linear rate can be maintained by retraining methods in the presence of weak performative shifts~\citep{perdomo2020performative}. 

There are various works in the optimization literature that aim to relax the strong-convexity condition on the objective function while maintaining favorable convergence properties. Examples include error bounds (EB)~\citep{luo1993error, drusvyatskiy2018error}, essential strong convexity (ESC)~\citep{stromberg2011duality, liu2014asynchronous}, weak strong-convexity (WSC)~\citep{ma2016linear, necoara2019linear}, restricted secant inequality (RSI)~\citep{zhang2013gradient, zhang2017restricted}, restricted strong-convexity (RSC)~\citep{negahban2012restricted, zhang2013gradient}, Polyak-Lojasiewicz (PL)
inequality~\citep{polyak1963gradient, loizou2021stochastic} and quadratic growth (QG) condition~\citep{anitescu2000degenerate, drusvyatskiy2018error}. The relations between these conditions have been studied in~\citep{liu2015asynchronous, karimi2016linear, bolte2017error, zhang2017restricted}.
In particular, for a smooth function with Lipschitz-continuous gradients, ~\citet{karimi2016linear} showed the following chain of implications:
\begin{equation}(SC) \xrightarrow[]{} (ESC) \xrightarrow[]{} (WSC) \xrightarrow[]{} (RSI) \xrightarrow{} (EB) \equiv (PL) \xrightarrow[]{} (QG).
     \label{eq:chain}
 \end{equation}
This implication chain is widely used in numerical optimization~\citep{wang2018differentially, nouiehed2019solving, vaswani2019fast}. 

We build on this work to analyze performative prediction from an optimization perspective. Previous works assumed the loss $\ell$ satisfies strong convexity (SC) globally which is the strongest condition according to the implication chain in Eq.~\ref{eq:chain}. Our contribution is to show that weaker conditions on $\DPR$ still imply desirable local properties of $\PR$. Specifically, in this work, we prove that assuming WSC and RSI of $\DPR$ is sufficient to establish WC and RSI of $\PR$, respectively, following similar steps taken in~\citet{miller2021outside}. These local conditions only need to hold around the minimizer and do not posit strong structural restrictions on the overall landscape such as SC, thus making the assumptions more realistic in practice. 

% Besides, we also discuss what conditions are presented for $\PR$ when the loss function satisfies the PL condition, which is among the weakest assumptions according to the implication chain and has been shown to be ubiquitous in over-parameterized neural networks~\citep{zhou2021local, liu2022loss}.
% \textcolor{red}{the Last sentence is very unclear, but we should first agree on what to say about PL.}

\subsection{Establishing weak convexity of the performative risk}

% Specifically, we demonstrate one can relax the strong convexity assumption on the loss function to two weaker conditions: WSC and RSI. We start by assuming conditions on $\DPR$ that only hold locally and derive local properties of $\PR$ following similar steps taken in~\citet{miller2021outside}. In contrast to WSC and RSI which have first-order dependencies on the gradient $\nabla \PR(\theta)$, the PL condition has a more complex dependency in terms of $\|\nabla \PR(\theta)\|^2$ which makes the analysis more complicated. 

% Careful validations for conditions required for WC and RSI, like how reliable and practical they are, are also needed. We remark on special cases for PL conditions and call for more focus on this topic.

First, we introduce the \emph{weak strong convexity}~\citep{karimi2016linear, necoara2019linear} in order to relax the strong convexity conditions. We will follow the convention that  $x_p$ denotes the projection of $x$ onto the optimal solution set $\gX^*$ throughout this paper.
\begin{definition}[WSC]
A function $f: \sR^d \to \sR$ is $\mu$-WSC if for all $x$ we have
\begin{equation}
    f^* \ge f(x) + \langle \nabla f(x), x_p - x \rangle + \frac{\mu}{2} \|x_p - x\|^2.
\end{equation}
where $x_p \in \gX^* = \arg \min_{x} f(x)$ and $f^*=f(x_p)$. 
\end{definition}

We translate this definition to the decoupled performative risk as follows:
\begin{assumption}
\label{assump_wsc}
For performative optimum $\thetaPO$ and its induced distribution $\gD=\gD(\thetaPO)$, suppose the optimal solution for minimizing $ \DPR(\thetaPO, \cdot)$ is $\theta^*$. We say $\DPR(\thetaPO, \cdot)$ satisfies $\mu-$WSC, if for any $\theta \in \Theta$ it holds that
\begin{align} 
     \DPR(\thetaPO, \theta^*) \ge \DPR(\thetaPO, \theta) + \nabla_\theta \DPR(\thetaPO, \theta)^\top (\theta^* - \theta) + \frac{\mu}{2} \|\theta^* - \theta\|^2.\label{eq_wsc}
\end{align}
\end{assumption}

This definition requires that the minimizer for the expected loss on the distribution
$\gD$ is unique: $
    \theta^* = \argmin_{\theta \in \Theta} \DPR(\thetaPO, \theta).
$ This is a mild assumption used by previous work focusing on repeated optimization in performative environments~\citep{mendler2020stochastic, drusvyatskiy2020stochastic, wood2021online}.
Intuitively, we posit a local property on the structure of $\DPR$ over the \emph{specific} distribution $\gD$ induced by $\thetaPO$. The quantity is measured over the domain of model parameter $\theta$.
We note that this condition is already weaker than the strong convexity Assumption~\ref{assump_strong_cvx} which is assumed to be true for \emph{any data point $z$}~\citep{miller2021outside}, thus depending heavily on the structure of the loss function.

Equipped with Assumption~\ref{assump_wsc}, we are ready to state the following new theorem.

\begin{theorem}\label{thm_wsc}
Under Assumptions~\ref{assump_smooth}, \ref{assump_sensitive}-\ref{assump_wsc}, suppose the performative optimum is also performative stable, then the
performative risk is guaranteed to be weakly convex if $\frac{\mu}{2\beta}\ge \epsilon$.
\end{theorem}
% \cnote{the assumption stable=optimal could need some more discussion and justification. 1) One could argue that you talk about how hard it is to find optimal points and then you make this assumption that somewhat removes the need to optimize through the distribution. retraining could again be used to find these points because they are stable. Just anticipating some questions }
We make justification on the assumption that $\thetaPO$ is also performative stable. As mentioned, optimal points and stable points do not generally imply each other. Therefore, simply adopting retraining (e.g., ~\citep{perdomo2020performative}) does not return global optima.  Nevertheless, the local properties of $\PR(\theta)$ under this assumption would be really helpful in determining the optimality of a performative stable point. 
In the next section, we will extend to more results on what information about $\thetaPO$ we could obtain at a specific stable point without the assumption.

\begin{proof}[Proof sketch]
Technically, our derivation starts from Eq.~\ref{eq_wsc}. After expanding the inequality around the minimizer of $\thetaPO$, we split $\nabla \PR$ into two terms: gradient of the loss function $\E_{z \sim \gD(\theta)} \nabla_\theta \ell(z; \theta)$, and gradient on the variable distribution $\E_{z \sim \gD(\theta)} \ell(z; \theta) \nabla_\theta \log{p_\theta(z)}$, where $p_\theta$ is the probability distribution function of $\gD(\theta)$. We then use the sensitivity Assumption~\ref{assump_sensitive} to bound the first term by the quadratic distance $\|\thetaPO-\theta\|^2$ and use the mixture dominance Assumption~\ref{assump_mix} to control the second term.
\end{proof}

The $\frac{\mu}{2\beta}$ term in the theorem is a sharp threshold for weak convexity of the performative risk, which aligns with the observations in~\citep{perdomo2020performative, miller2021outside}. See details in Appendix~\ref{App:ext}.
\subsection{Establishing RSI of the performative risk}
The second condition we consider is that \emph{Restricted Secant Inequality}~\citep{zhang2013gradient}. It is defined as follows:
\begin{definition}[RSI]
A function $f: \sR^d \to \sR$ satisfies Restricted Secant Inequality (RSI) if for all $x$ we have
\begin{equation}
    \langle \nabla f(x), x - x_p \rangle \ge \mu \|x_p - x\|^2.
\end{equation}
Again, $x_p$ is
the projection of $x$ onto the optimal solution set $ \gX^*$. 
\end{definition}

\begin{remark}
A function satisfies restricted strongly convex (RSC) if it is convex and also satisfies RSI.
\end{remark}

\begin{assumption}
\label{assump_rsi}
For performative optimum $\thetaPO$ and its induced distribution $\gD$, suppose the optimal solution for minimizing $ \DPR(\thetaPO, \cdot)$ is unique and is denoted as $\theta^*$. We say $\DPR(\thetaPO, \cdot)$ satisfies $\mu-$RSI, if for any $\theta$ it holds
\begin{align} 
    \langle \nabla_\theta \DPR(\thetaPO, \theta), \theta - \theta^* \rangle \ge \mu \|\theta^* - \theta\|^2
\end{align}
\end{assumption}
\begin{theorem}\label{thm_rsi}
Under Assumptions~\ref{assump_lipschitz}-\ref{assump_mix}, and ~\ref{assump_rsi}, suppose the performative optima is also performative stable, then $\PR(\theta)$ satisfies RSI.
\end{theorem}
Note that compared to WSC, the RSI condition is weaker and thus generalizes more easily in first-order optimization. See details in Appendix~\ref{App:ext}. We conjecture that specific implementations discussed by~\citet{negahban2012restricted, zhang2013gradient, zhang2017restricted} could be extended to optimize the $\PR$ in this setting.

\section{When are local properties sufficient for stability and optimality?}
\label{sec:compare}

As aforementioned, existing methods in the literature that seek stable points do not have global optimality guarantees. Meanwhile, current methods aiming at optimizing $\PR$ \emph{directly} require local structural assumptions around the performative optima that may not hold in real settings. 

In this section, we investigate relations between stable points and optimal points. We relate to previous work focusing on the convergence of retraining methods to stable points~\citep{mendler2020stochastic, drusvyatskiy2020stochastic,perdomo2020performative, wood2021online, bianchin2021online} and argue that if  stable points offer a good approximation for optimal points, then they can serve as a good starting point to ensure convergence to optimal points with only local assumption on the performative risk.

% That being said, we could first arrive at stable points, then start optimization and get convergence guarantees because we initialize within the locally convex region. Intuitively, we work on the \emph{trust region} for stable points in performative prediction schemework.

% Practically, we aim to distinguish a smaller set of $\theta \in \Theta$ that provable presents higher $\PR$ than current standing point $\theta_0$, thus one could turn to optimize $\PR$ in complement space for efficiency.

Suppose the Lipschitzness condition for loss (c.f. Assumption~\ref{assump_lipschitz}) holds throughout this section, which is standard and mild in learning theory or optimization theory literature. For any $\theta,\theta^\prime$ in the parameter domain, we could quantify the gap between $\PR(\theta)$ and $\PR(\theta^\prime)$ as
\begin{align*}
    \PR(\theta^\prime)&= \DPR(\theta^\prime,\theta^\prime) \\
    &\geq \DPR(\theta,\theta^\prime) - L W(\gD(\theta),\gD(\theta^\prime)) \\
    &= \PR(\theta) + [\DPR(\theta,\theta^\prime)-\DPR(\theta,\theta)] - L W(\gD(\theta),\gD(\theta^\prime)) .
\end{align*}

% \begin{proposition}
% Under Assumption~\ref{assump_lipschitz}, $\theta$ is performative optimal if 
% \[W(\gD(\theta),\gD(\theta^\prime))\leq \frac 1 L [\DPR(\theta,\theta^\prime)-\PR(\theta)]\]
% for every $\theta^\prime\in\Theta$.
% \end{proposition}

The second term on the RHS measures the distance in the function value between $\theta$ and $\theta^\prime$ over the distribution $\gD(\theta)$. 

In the remainder of this section, we pick  $\theta$ to be a stable point (i.e., $\theta = \arg \min_{x} \DPR(\theta, x)$) so we can work with the suboptimality gap of the function. First rewrite the above equation as 
\begin{align}
    \PR(\theta^\prime)
    &\geq \PR(\theta) + \Delta_\theta(\theta^\prime)- L W(\gD(\theta),\gD(\theta^\prime)).
\label{eq:opt_condition}
\end{align}
where $\Delta_\theta(\theta^\prime):= \DPR(\theta, \theta^\prime) - \PR(\theta) \ge 0$ is the suboptimality of $\theta^\prime$ measured over $\gD(\theta)$ 

\begin{proposition}
\label{prop}
Under Assumption~\ref{assump_lipschitz}, if
\begin{equation} 
\Delta_\theta(\theta^\prime)\geq L W(\gD(\theta),\gD(\theta^\prime)),
\label{eq:cond}
\end{equation}
it holds that 
$$\PR(\theta)\leq \PR(\theta^\prime).$$
\end{proposition}

The proof is straightforward by noticing that, when Eq.~\ref{eq:cond} holds, we have $\PR(\theta) \le \PR(\theta^\prime)$ for all $\theta^\prime \in \Theta$. Thus $\theta$ is performative optimal according to definition. Intuitively, Eq.~\ref{eq:cond} relates changes in the loss to changes in the distribution. 

\paragraph{Starting from $\theta$.} Suppose we have reached $\theta$ which is a performative stable point. To achieve $\thetaPO$, we may consider two directions. (1) How is the performance of $\PR(\theta)$? What is the gap between $\PR(\theta)$ and the optimal performative risk (denoted as $\PR^*$)? 
(2) What geometric information about $\thetaPO$ could we obtain at $\theta$? As we may believe that $\theta$ will serve as a good approximation for $\thetaPO$ when the Euclidean distance is close.
We believe investigating along the two directions is promising since it makes use of the large amount of work focusing on reaching $\thetaPS$. 

Below, we give three preliminary examples on how some general structural conditions on the loss function and distribution maps could lead to useful results related to the two directions.
% According to Proposition~\ref{prop}, we only need Eq.~\ref{eq:cond} to show the optimality of $\theta$. So the problems turn into: what natural conditions on the loss (e.g. convexity, PL, etc.) and conditions on the distribution map are needed to imply Eq.~\ref{eq:cond}.
 
%  Below we show that (a type of boundedness or Lipschitzness)

We show the suboptimality of $\theta$ under Lipschitzness and a type of boundedness.
\begin{example}
Assume performative shifts are bounded by an absolute value, i.e., $$
W(\gD(\theta),\gD(\theta^\prime))\leq B.
$$
We have the following bound characterizing the suboptimality of $\theta$
\begin{align}
    \PR(\theta)-\PR^*
    &\leq L B.
\end{align}
\end{example}
\begin{proof}
The proof is straightforward by noticing  $\Delta_\theta(\theta^\prime) \ge 0$ when $\theta$ being stable. From Eq.~\ref{eq:opt_condition} we have $\PR(\theta^\prime) \ge \PR(\theta) - LB $.
\end{proof}

The next two examples will show that $\thetaPO$ could not be far from $\theta$. 
\begin{example}
Assume (1) performative shifts are bounded by an absolute value $B$ and (2) $\Delta_\theta(\theta^\prime)$ satisfies quadratic growth, i.e.,  $ \Delta_\theta (\theta^\prime) \ge \gamma\|\theta-\theta^\prime\|^2 $, we have that performative optimal point $\thetaPO$ satisfies
$$
\|\thetaPO - \theta\| \le \sqrt{\frac{LB}{\gamma}}.
$$
\end{example}
\begin{proof}
We proof by contradiction. Suppose there exists a performative optimal point $\thetaPO$ which is $\sqrt{\nicefrac{LB}{\gamma}}-$close to $\theta$. The quadratic growth shows $$\Delta_\theta(\thetaPO)\ge LB \ge L W(\gD(\theta), \gD(\thetaPO)).$$
From Eq.~\ref{eq:opt_condition} we have $\PR(\thetaPO) \ge \PR(\theta)$ which contradicts with the optimality of $\thetaPO$.
\end{proof}

% \textcolor{red}{So here I get $\PR(\theta)-\PR^* \leq L B ( \gamma + 1)$. Is this what you get as well? Can you add the steps of your derivation?}

\begin{example}Under Assumption~\ref{assump_sensitive}, suppose $\Delta_\theta(\theta^\prime)$ satisfies quadratic growth, the performative optimal point $\thetaPO$ satisfies
$$
\|\thetaPO - \theta\| \le \frac{L \epsilon}{\gamma}.
$$
\end{example}
\begin{proof}
We proof by contradiction. Suppose there exists a performative optimal point $\thetaPO$ which is $\nicefrac{L\epsilon}{\gamma}-$close to $\theta$. The quadratic growth shows $\Delta_\theta(\thetaPO)\ge \frac{L^2 \epsilon^2}{\gamma}$. At the meantime, senstitivity assumption~\ref{assump_sensitive} shows 
$$
W(\gD(\theta), \gD(\thetaPO)) \le \epsilon \|\theta - \thetaPO\| \le \frac{L \epsilon^2}{\gamma}.
$$
Therefore we have $\Delta_\theta(\thetaPO) \ge L W(\gD(\theta), \gD(\thetaPO))$.  Substituting the relation into Eq.~\ref{eq:opt_condition} shows $\PR(\thetaPO) \ge \PR(\theta)$ which contradicts with the optimality of $\thetaPO$.
\end{proof}

% \cnote{add a summary here of what we learnt in this section to make it more digestable for the reader}
To conclude, in this section we attempt to make use of stable points in finding optima points. Though $\thetaPS$ and $\thetaPO$ do not imply each other as discussed aforementioned, we aim at revealing when and how stable points could be served as starting points for finding optima points. In such scenario, previous work focusing on convergence of retraining methods to stable points would still be worthy.

Concretely, we have shown several conditions under which stable points offer a good approximation for optimal points. Then, with only local assumption on $\PR$, we show that one could start from stable points in an optimization procedure, and finally, converge to optimal points.

\begin{remark}
An interesting special case where stable points are optimal without requiring some sort of global optimality is when the Bayesian error of every distribution is the same. This means all minima are on one level-set. Hence, no other point can have a smaller loss.
\end{remark}

% \textcolor{red}{al: I'm missing what the take-home message of this section is. It feels rather disorganized without a clear thread. I would try to think about what are one or two important messages and rewrite this section around these messages. For instance, the conditions in Propositions 1 and 2 seem to be the most important insight, right? Then I would say that early on: i.e. we introduce two new conditions and we discuss their meaning on concrete examples. }

\section{Conclusion}
This paper studies optimization aspects of the performative risk minimization problem. 

First, we relax the standard strong convexity assumption on the loss function required in prior work to show convergence of iterative optimization methods in performative prediction. In particular, we study two weaker conditions on $\DPR$: WSC and RSI -- both are sufficient to establish weak convexity of the performative risk under suitable assumptions on the distribution map. Our work takes a first step towards importing advanced assumptions from the classical optimization literature into performative prediction.

Second, we make a contribution by raising interesting questions about the meaning of local regularity assumptions in the context of performative prediction. If stable points can be found heuristically and serve as a good approximation for global optima, the local regularity of $\PR$ can be sufficient to find optimal points. We provide some bounds on the distance between optimal and stable points but it remains to better understand how they relate in practical settings.

As of future work, we note that a very interesting and important topic to study is the PL condition which is among the weakest assumptions according to the implication chain and is ubiquitous in over-parameterized neural networks~\citep{zhou2021local}.

Suppose the following $\mu$-PL condition holds
 \[\frac{1}{2} \left\| \nabla_{\theta^\prime} \DPR(\theta, \theta^\prime) \right\|^2 \ge \mu (\DPR(\theta, \theta^\prime)- \PR(\theta)) = \mu \Delta_\theta (\theta^\prime).\]

Naturally, we expand the RHS through Eq.~\ref{eq:opt_condition}
\begin{align*}
    \frac{\left\| \nabla_{\theta^\prime} \DPR(\theta, \theta^\prime) \right\|^2}{2\mu} &\ge \PR(\theta^\prime) - \PR(\theta) - L W(\gD(\theta), \gD(\theta^\prime));.
\end{align*}

Intuitively, we raise several interesting questions related to finding performative optima.
\begin{enumerate}
    \item Understanding when and how (e.g., some structural properties of loss function or a natural set of distributions), it holds that $W(\gD(\theta), \gD(\theta^\prime)) \le C \| \nabla_{\theta^\prime} \DPR(\theta, \theta^\prime)\|^2 $ where $C$ is a certain constant. Since the condition characterizes local properties of $\DPR$ near performative stable points, it could be more common in practice. Notice that current performative prediction literature heavily rely on sensitivity relation (c.f. Assumption~\ref{assump_sensitive}). 
    \item What is the impact of data pre-processing steps on the implications of performative shifts? Can they influence performativity effects? For example, can data whitening, result in distribution with favorable properties, such as location-scale family?
\end{enumerate}

\section*{Acknowledgements}
The author gratefully acknowledges Aurelien Lucchi and Celestine Mendler-Dünner for numerous helpful discussions and advice during the internship at ETH Zürich.

\bibliography{references}

\newpage
\clearpage
\appendix
\label{sec:app}
\section{Proofs for Section~\ref{sec:ext}} \label{App:ext}

[Proofs for Theorem~\ref{thm_wsc}]
\begin{proof}
At $\thetaPO$, since performative optimum is also performative stable, it is known
$\theta^* = \thetaPO$.

Using WSC property we have
\begin{equation}
\E_{z \sim \gD(\thetaPO)} \ell(z; \thetaPO) \ge \E_{z \sim \gD(\thetaPO)} \ell(z; \theta) + \langle \nabla_\theta \DPR(\thetaPO, \theta), \thetaPO - \theta \rangle + \frac{\mu}{2} \| \thetaPO - \theta\|^2.
\end{equation}
Our goal is to characterize the function class of $\PR(\theta) = \E_{z \sim \gD(\theta)} [l(z; \theta)]$

Potentially, we expect it to be weak convex w.r.t. $\theta$, i.e.
\begin{equation} \label{eq_goal_1}
\PR(\thetaPO) \ge \PR(\theta) + \langle \nabla \PR(\theta), \thetaPO - \theta \rangle,
\end{equation}
or using the definition of $\PR(\theta)$:
\begin{equation}
\E_{z \sim \gD(\thetaPO)} [l(z; \thetaPO)] \ge \E_{z \sim \gD(\theta)} [l(z; \theta)] + \langle \nabla \E_{z \sim \gD(\theta)} [l(z; \theta)], \thetaPO - \theta \rangle.
\end{equation}

First, we separate the gradient of $\PR$ into
\begin{align}
    \nabla_\theta \PR(\theta) = \overbrace{\E_{z \sim \gD(\theta)} \nabla_\theta \ell(z; \theta)}^{\nabla_1} + \overbrace{\E_{z \sim \gD(\theta)} \ell(z; \theta) \nabla_\theta \log{p_\theta(z)} }^{\nabla_2}.
\end{align}

We have
\begin{align*}
    & \qquad \PR(\thetaPO)  \ge \PR(\theta) + \langle \nabla \PR(\theta), \thetaPO - \theta \rangle\\
    &\stackrel{(i)}{\Leftarrow}  \DPR(\thetaPO, \theta) \ge \PR(\theta) + \nabla^\top (\thetaPO - \theta) - \langle \E_{z \sim \gD(\thetaPO)} \nabla \ell(z; \theta), \thetaPO - \theta \rangle - \frac{\mu}{2} \|\thetaPO -\theta\|^2\\
    &\Leftrightarrow  \DPR(\thetaPO, \theta) \ge \PR(\theta) + \nabla_2^\top (\thetaPO- \theta) + \langle \nabla_1 - \E_{z \sim \gD(\thetaPO)} \nabla \ell(z; \theta), \thetaPO - \theta \rangle - \frac{\mu}{2} \|\thetaPO -\theta\|^2\\
    &\stackrel{(ii)}{\Leftarrow}  \DPR(\thetaPO, \theta) \ge \PR(\theta) + \nabla_2^\top (\thetaPO- \theta) + (\beta \epsilon - \frac{\mu}{2}) \|\thetaPO -\theta\|^2\\
    &\stackrel{(iii)}{\Leftarrow}  \frac{\mu}{2} \ge \beta \epsilon
\end{align*}

\begin{enumerate}[(i).]
    \item Here we use: $\PR(\thetaPO) \ge \DPR(\thetaPO, \theta) + \langle \E_{z \sim \gD(\thetaPO)} \nabla \ell(z; \theta), \thetaPO - \theta \rangle + \frac{\mu}{2} \|\thetaPO - \theta\|^2$.
    \item This is because $\epsilon$-sensitive of distribution map $\gD$.
    \item Mixture dominance implies: $\DPR(\thetaPO, \theta) \ge \PR(\theta) + \nabla_2^\top (\thetaPO-\theta)$.
\end{enumerate}

We conclude that we only need weak strong convexity of $\DPR$, local mixture dominance and local distribution sensitivity to guarantee that $\PR$ is weakly-convex to $\thetaPO$, instead of strongly convex everywhere.
\end{proof}

[Proofs for Theorem~\ref{thm_rsi}]
\begin{proof}
At $\thetaPO$, because performative optimum is also performative stable, it is known $\theta^* = \thetaPO$. 

Using RSI property we have
$$
\langle \nabla_\theta \DPR(\thetaPO, \theta), \theta - \thetaPO \rangle \ge \mu \|\thetaPO -\theta\|^2.
$$

We derive back as following, we want to show $\PR$ is $RSI$, i.e.,
\begin{align*}
    & \qquad \langle \nabla_\theta \PR(\theta), \theta - \thetaPO \rangle \ge \mu^\prime \|\thetaPO -\theta\|^2\\
    & \Leftrightarrow 0 \ge \mu^\prime \|\thetaPO -\theta\|^2 + \langle \nabla_1, \thetaPO - \theta \rangle + \langle \nabla_2, \thetaPO - \theta \rangle\\
    & \Leftrightarrow 0 \ge \mu^\prime \|\thetaPO -\theta\|^2 + \langle \nabla_1 - \nabla_\theta \DPR(\thetaPO, \theta), \thetaPO - \theta \rangle + \langle \nabla_\theta \DPR(\thetaPO, \theta), \thetaPO-\theta \rangle +\langle \nabla_2, \thetaPO - \theta \rangle\\
    & \stackrel{(i)}{\xLeftarrow{}} 0 \ge (\mu^\prime + \beta \epsilon) \|\thetaPO -\theta\|^2 + \langle \nabla_\theta \DPR(\thetaPO, \theta), \thetaPO-\theta \rangle + \langle \nabla_2, \thetaPO - \theta \rangle\\
    & \stackrel{(ii)}{\xLeftarrow{}} 0 \ge (\mu^\prime + \beta \epsilon) \|\thetaPO -\theta\|^2 + \langle \nabla \DPR(\thetaPO, \theta), \thetaPO - \theta \rangle + \DPR(\thetaPO, \theta) - \PR(\theta) \\
    & \stackrel{(iii)}{\xLeftarrow{}} 0 \ge (\mu^\prime + \beta \epsilon + L \epsilon) \max(\|\thetaPO -\theta\|, \|\thetaPO -\theta\|^2) + \langle \nabla \DPR(\thetaPO, \theta), \thetaPO - \theta \rangle
\end{align*}
because
\begin{enumerate}[(i).]
    \item Because we have $\epsilon$-sensitiveness of distribution map $\gD$.
    \item From mixture dominance we know $\DPR(\thetaPO, \theta) \ge \PR(\theta) + \langle \nabla_2, \thetaPO - \theta \rangle$.
    \item This is because of  $L$-Lipschitzness of loss function w.r.t. $z$.
\end{enumerate}

We have
1) If $\|\thetaPO -\theta\|^2 \leq \|\thetaPO -\theta\|$ (close to optimum), then 
\begin{align}
 \underbrace{\langle \nabla \DPR(\thetaPO, \theta), \theta - \thetaPO \rangle}_{\ge \mu \|\thetaPO -\theta\|^2} \ge (\mu^\prime + \beta \epsilon + L \epsilon) \|\thetaPO -\theta\|
\end{align}
The retrodiction shows when assuming $\DPR$ is $\mu$-RSI, then our goal is achieved if
\begin{align}
    \mu^\prime = \mu \|\thetaPO  - \theta\| - (\beta + L) \epsilon \ge 0
\end{align}
2) If $\|\thetaPO -\theta\| \leq \|\thetaPO -\theta\|^2$ then 
\begin{align}
 \langle \nabla \DPR(\thetaPO, \theta), \theta - \thetaPO \rangle \ge \mu \|\thetaPO -\theta\|^2 \ge (\mu^\prime + \beta \epsilon + L \epsilon) \|\thetaPO -\theta\|^2
\end{align}
The retrodiction shows: when assuming $\DPR$ is $\mu$-RSI, then our goal is achieved if
\begin{align}
    \mu^\prime = \mu - (\beta + L) \epsilon \ge 0
\end{align}
Proof is completed.
\end{proof}

\end{document}